\documentclass[review]{elsarticle}
\usepackage{lineno,hyperref}
\modulolinenumbers[5]

\usepackage[ruled, linesnumbered,noend]{algorithm2e}
\newtheorem{definition}{Definition}
\newtheorem{lemme}{Lemma}
\newtheorem{propriete}{Property}
\newtheorem{proposition}{Proposition}
\newtheorem{preuve}{Proof}
\newtheorem{example}{Example}
\usepackage{multirow}
\usepackage{colortbl}
\usepackage{xcolor}
\usepackage{inferance}
\usepackage{deduc}
\usepackage{arydshln}
\usepackage{amsmath,amssymb,amsfonts}
\usepackage{arydshln}
\usepackage{eurosym}

\begin{document}

\begin{frontmatter}

\title{Extracting Seasonal Gradual Patterns from Temporal Sequence Data Using Periodic Patterns Mining}


\author[mymainaddress]{Jerry Lonlac\corref{mycorrespondingauthor}}
\cortext[mycorrespondingauthor]{Corresponding author}
\ead{jerry.lonlac@imt-lille-douai.fr}

\author[mymainaddress]{Arnaud Doniec}
\author[mymainaddress]{Marin Lujak}
\author[mymainaddress]{Stephane Lecoeuche}

\address[mymainaddress]{IMT Lille Douai, Institut Mines-T\'el\'ecom,   Univ. Lille, Center for Digital Systems, F-59000 Lille, France}

\begin{abstract}
Mining frequent episodes aims at recovering sequential patterns from temporal data sequences, which can then be used to predict the occurrence of related events in advance.
On the other hand, gradual patterns that capture co-variation of complex attributes in the form of ``when X increases/decreases,
Y increases/decreases'' play an important role in many real world applications where huge volumes of complex numerical data must be handled.
Recently, these patterns have received attention from the data mining community exploring temporal data who proposed methods to automatically extract gradual patterns from temporal data.
However, to the best of our knowledge, no method has been proposed to extract gradual patterns that regularly appear at  identical time intervals  in many sequences of temporal data, despite the fact that such patterns may add knowledge to certain applications, such as e-commerce.
In this paper, we propose to extract  co-variations of periodically repeating attributes from the sequences of temporal data that we call \textit{seasonal gradual patterns}.
For this purpose, we formulate the task of mining seasonal gradual patterns as the problem of mining periodic patterns in multiple sequences and then we exploit periodic pattern mining algorithms to extract \textit{seasonal gradual patterns}.
We discuss  specific features of these patterns and propose an approach for their extraction based on mining periodic frequent patterns common to multiple sequences. 
We also propose a new anti-monotonous support definition associated to these seasonal gradual patterns.
The illustrative results obtained from some real world data sets 
show that the  proposed approach is efficient  and that it can extract  small sets of patterns by filtering numerous nonseasonal patterns to identify the seasonal ones.
\end{abstract}


\end{frontmatter}


\section{Introduction}
\label{sec:intro}

Due to the abundance of data collection devices and sensors, numerical data are ubiquitous and produced in increasing quantities. They are produced in many domains including e-commerce, biology, medicine, environment and ecology, telecommunications, and system supervision. In recent years, the analysis of numerical data has received attention from the data mining community and methods have been defined for dealing with such data.
These methods have allowed to automatically extract different kinds of knowledge from numerical data express under the form of patterns such as quantitative itemset/association rules \cite{Srikant1996J,AumannL99,Salleb-AouissiVN07} and interval patterns \cite{KaytoueKN11}.
Recently, gradual patterns that model frequent co-variations between numerical attributes, such as ``the more experience, the higher the salary, and the lower the free time'' aroused great interest in a multitude of areas. For example, in medicine,  gradual patterns make it possible to capture the correlations between memory and feeling points from the \textit{Diagnostic and statistical manual of mental disorders} \cite{american2013diagnostic}; in biology, they help to analyze genome data for discovering correlations between genomic expressions \cite{AgierPS07}; in health, where researchers often seek to explain  the differences between micro sequences (DNA, gene or protein) according to   clinical characteristics (grade of a tumor, number of recurrences, etc.) \cite{BringayLOST09}; in financial markets, where one would like to discover co-evolution between  financial indicators, or in marketing for analyzing client databases \cite{DoTLNTA15}.
Several works have addressed the mining of gradual patterns and different algorithms have been designed for discovering gradual patterns from different numerical data models (e.g., temporal, stream, relational, or noisy data) 
\cite{DoTLNTA15,Hullermeier02,BerzalCSVS07,OudniLR13,Di-JorioLT09,NinLP10,NegrevergneTRM14,PhanIMPT15}.

Although certain  state-of-the art algorithms (e.g., \cite{Masseglia08gradualtrends,Lonlac18,OwuorLO19}) deal with temporally annotated numerical data, they can be only used on the data constituted as a single valued sequence.
When considering the databases constituted as many ordered value sequences, state-of-the art gradual pattern mining algorithms do not allow extracting any other kind of knowledge that is typical of sequential data.
While seasonal gradual patterns could make it possible to discover  seasonality  from the graduality point of view in the temporal sequence data (gradual patterns that regularly appear at   identical time intervals). Such patterns are suitable for recommending items to be supplied in  retail stores (at each period) by targeting the groups of items based on  common variations of their stock quantity.

Let us consider an inventory database in a retail store in which the stock quantity of each product (item) is indicated for each day or month. Suppose that this database contains items $i_1$ and $i_2$.
Furthermore, let us assume that the quantities of items $i_1$ and $i_2$ co-vary regularly at identical time intervals.
State-of-the-art algorithms for mining gradual patterns would fail to discover this co-variation between $i_1$ and $i_2$ for the following reasons. First, to find frequent gradual patterns, they treat a database with multiple sequences as a single sequence and are thus unable to consider seasonality. Second, the graduality between items $i_1$ and $i_2$ can appear in the seasonality of different sequences (i.e., the quantity of item $i_1$ increases/decreases in  season $T$
of sequence $s_1$ and the quantity of item $i_2$ increases/decreases in season $T$  of sequence $s_2$).

To find these patterns from temporal sequence data, a naive solution is to apply any of the state-of-the art algorithms described above on each sequence to extract classical gradual patterns with the associated graduality periods (set of sequence  objects called extensions, supporting the gradual pattern \cite{JabbourLS19}), then to combine the patterns found in each sequence to find gradual patterns common to multiple sequences and associated with the same extensions. However, combining these patterns would be very costly due to a very high quantity of gradual patterns and the extensions associated to each pattern generated from  each sequence; it would be inefficient to keep them in memory.
In this paper, we propose  to extract seasonal gradual patterns from temporal sequence data by exploiting the  algorithms for mining periodic patterns  common to multiple sequences in order to avoid the problem of storing huge amounts of patterns and to overcome the issues  mentioned above.

Periodic patterns that capture repeated regularities in  data sequences have attracted considerable attention recently. They are often sought in numerous application domains, for example in biology for detecting sets of repeating DNA molecules which is necessary to find what leads to some external expressions and in marketing to identify products that are regularly bought by some customers (e.g. every week or month) to promote the sales of groups of items.
Periodic pattern mining algorithms are generally used to mine such patterns in a  single sequence \cite{Fournier-VigerL16,TanbeerAJL09,GalbrunCTTC18} or in sets of sequences (a sequence database) \cite{Fournier-VigerL19} for discovering periodic patterns common to several sequences. The latter case is suitable to discover periodic behavior common to several customers from a customer sequence database. For instance, for market basket analysis, periodic patterns can be used to find that many customers have the same sequential behavior over time and to improve sales and marketing strategies. There are many other applications where periodic patterns have a significant interest as web usage recommendation \cite{Fong11} or weather prediction.
We exploit periodic patterns mining algorithms common to multiple sequences for extracting seasonal gradual patterns (the latter term proposed by \cite{Lonlac18}), but regularly associated to the same sequence of objects in all value sequences. We justify the interest for such patterns in the e-commerce and paleoecology\footnote{the study of interactions between organisms and/or interactions between organisms and their environments across geologic timescales} domains where it is very important to understand seasonal patterns. 
In fact, in the  e-commerce domain, the knowledge brought by such patterns can be used  in logistics for decision marking in, e.g.,  inventory and supply chain management.
In the paleoecological domain, the proposed seasonal gradual patterns are interesting for discovering co-evolution groupings of multi-varied paleoecological indicators common to several environments (the interest of gradual patterns respecting a temporal order has been shown in \cite{Lonlac18}).
Our approach extends gradual patterns in the temporal context such that we include the seasonal correlations between attributes.

The main contributions of this paper are summarized as follows:
\begin{itemize}
    \item We propose the problem of mining seasonal gradual patterns (namely  attribute co-variations that appear regularly in multiple sequences at identical time sub-sequences) in a temporal sequence data. We then study the properties of this problem.

    \item We propose a new support definition of gradual patterns for evaluating if gradual patterns are seasonal (or whether they appear regularly at   identical time intervals  in multiple sequences of temporal data).

    \item We propose an algorithm named \textit{MSGP (Mining Seasonal Gradual Patterns)} to efficiently find all seasonal gradual patterns. The algorithm relies on: i) the transformation of the initial temporal sequence numerical data into a database with a sequence of items and ii) the use of a modified  version of \textit{MPFPS\_BFS} (Mining Periodic Frequent Pattern common to multiple Sequences)  algorithm \cite{Fournier-VigerL19}.

    \item We conduct multiple experiments on three real world databases in order to validate this proposition and to compare its results to that of the benchmark algorithms in terms of scalability and runtime. The experiment results reveal interesting patterns and show that the  proposed \textit{MSGP} algorithm is able to efficiently discover all seasonal gradual patterns from the temporal sequence data.
\end{itemize}

The rest of the paper is organized as follows. We present in section \ref{relatedWork} related works. Section \ref{pre} introduces  preliminary notions on the mining of gradual and periodic patterns. Section \ref{problemStement}  defines the problem of mining seasonal gradual patterns. Section \ref{SeasonalPatternsExtraction} presents the proposed \textit{MSGP} algorithm. Before concluding in section \ref{conclusion}, we present  the experimental evaluation and the obtained results in Section \ref{Experiments}.

\section{Related work}
\label{relatedWork}

Discovering frequent simultaneous attribute co-variations (gradual patterns) from numerical databases is a  more complex task than the mining of extract classical itemsets. The former  computes all  possible orders to find the most representative ordering of all the database objects.
Gradual patterns have been extensively used in many real world applications due to their advantages in exploring numerical data. They allow to extract more expressive knowledge representing the variability of numerical values as well as their possible interdependencies.
Several works have addressed the gradual patterns mining problem and many efficient algorithms have been designed to automatically extract these patterns from a numerical database.
For example, an algorithm called GRITE (GRadual ITemset Extraction)  \cite{Di-JorioLT09} was proposed to automatically extract gradual patterns from large databases.
This algorithm takes advantage of a binary representation of a lattice structure that represents data by a graph whose nodes are defined as the objects of the numerical database and the links stand for the precedence relations derived from the attributes taken into account. The authors adopt a binary representation of the graph by a matrix and assess the frequency of a gradual pattern as the length of the longest path in the graph respecting this gradual pattern.
A drawback of the previous approach is that the number of gradual patterns extracted can be large, making their interpretation by an expert difficult or impossible. One solution to reduce the number of extracted patterns is to use   constraints in the mining process  to focus on patterns of interest.
Closed gradual patterns \cite{DoTLNTA15,NegrevergneTRM14} were proposed to overcome  this issue.
In fact, from a set of specific gradual patterns (closed gradual patterns), it is possible to regenerate the set of all gradual patterns. Moreover, with the closed patterns, redundant information is avoided.

According to the considered data model (temporal data, data stream, multi-relational data, temporal data, graph data, etc.) and the intended application, different algorithms were proposed in order to discover variants of gradual patterns in the numerical data supplied with specific constraints. for expressing another kind of knowledge. For example, on the multi-relational database, the notion of gradual patterns was extended to the case in which the co-variations are possibly expressed between attributes of different database relations \cite{PhanIMPT15}. So, a new class of gradual patterns called ``multi-relational gradual pattern'' was defined on the basis of both Kendall's and gradual supports and two algorithms were proposed for discovering them from a multi-relational database.
Gradual sequential patterns \cite{Masseglia08gradualtrends} were defined on the sequential data for discovering the trends in time-related databases based on fuzzy sequential pattern mining.
Several recent works dealt with finding other kinds of gradual patterns in the temporally annotated numerical data with related real world (rainfall and paleo-ecological) applications \cite{Lonlac18,OwuorLO19,lonlacIDA2020}.

Contrary to the described related work, in this paper, we are interested in mining   attribute co-variations  that appear  regularly  in multiple sequences  at   identical time  sub-sequences.

\section{Preliminary definitions}
\label{pre}
In this section, we formally describe the problem of mining frequent gradual itemsets (patterns) in a numerical database. Then, we present the problem of finding periodic patterns in multiple sequences (a sequence database) that we exploit  for our approach.

\subsection{Gradual patterns mining problem}
\label{sec:gradual}

The problem of mining gradual patterns consists in finding attribute co-variations in a numerical dataset of the form \textit{``The more/less X, . . . , the more/less Y''}. 
We assume herein that we are given a database $\Delta$ containing a set of observations $\mathcal{D}$ that defines a relation on an attribute set $\mathcal{I}$ with numerical values.
Let $d[i]$ denote the value of  attribute $i$ over observation $d$ for all $d \in \mathcal{D}$.

In Table \ref{tab:exampleDataset2}, we give an illustrative example of a numerical database built over the set of attributes $\mathcal{I} =$ $\{age, salary, cars, loans\}$.

\begin{table}
\centering
\begin{tabular}{ccccc}
 \hline
\textbf{tid} & \textbf{age} & \textbf{salary} & \textbf{cars} & \textbf{loans} \\
\hline
$d_1$ & 22 & 1000 & 2 &  4 \\
$d_2$ & 24 & 1200 & 3  & 3 \\
$d_3$ & 28 & 1850 & 2  & 5 \\
$d_4$ & 20 & 1250 & 4  & 2 \\
$d_5$ & 18 & 1100 & 4  & 2 \\
$d_6$ & 35 & 2200 & 4  & 2 \\
$d_7$ & 38 & 3200 & 1  & 1\\
$d_8$ & 44 & 3400 & 3  & 6\\
$d_9$ & 52 & 3800 & 3  & 3\\
$d_{10}$ & 41 & 5000 & 2  & 7\\
\hline
\end{tabular}
\caption{Database $\Delta_1$}
\label{tab:exampleDataset2}
\end{table}

Each attribute will hereafter be considered twice: once to indicate its increase and once to indicate its decrease, using the $\uparrow$ and $\downarrow$ variation symbols, where $\uparrow$ stands for increasing  and $\downarrow$ stands for decreasing variation.
In the following, for all $d, d' \in \mathcal{D}$ and for all $i \in \mathcal{I}$, we denote $i_{d,d'}^{\uparrow}$ (respectively $i_{d,d'}^{\downarrow}$) to mean that the value of attribute $i$ increases (respectively decreases) from $d$ to $d'$.

\begin{definition}[Gradual item]
\label{def:gradualItem}
Let $\Delta$ be a data set defined on a numerical attribute set $\mathcal{I}$. A gradual item is defined under the form $i^*$, where $i$ is an attribute of  $\mathcal{I}$ and $* \in \{\uparrow, \downarrow\}$. 
\end{definition}

If we consider the numerical database of Table \ref{tab:exampleDataset2}, $age^{\uparrow}$ (respectively $age^{\downarrow}$) is a gradual item meaning that the values of attribute \textit{age} are increasing (respectively decreasing).

A \textbf{gradual itemset (pattern)} $g = (i_1^{*_1}, ... , i_k^{*_k})$ is a non-empty set of gradual items. It imposes a variation constraint on several attributes simultaneously.
The length of a gradual itemset is equal to the number of gradual items that it contains.
A $k$-itemset is an itemset containing $k$ $(k > 1)$ gradual items.



For example, $g_1 = \{age^{\uparrow},  salary^{\uparrow}\}$ is a gradual pattern ($2$-itemsets) meaning that \textit{"the
higher the age, the higher the salary"}.


The support (frequency) of a gradual pattern amounts to the extent to which a gradual pattern is present in a given database. Several support definitions have been proposed in the literature (e.g., \cite{Hullermeier02,BerzalCSVS07,Di-JorioLT09,Lonlac18,CaldersGJ06}), showing that gradual patterns can follow different semantics.
Huellermeier  \cite{Hullermeier02} bases  the computation of the support of gradual patterns on linear regression.  Berzal et al. \cite{BerzalCSVS07}  and Calders et al. \cite{CaldersGJ06} consider the proportion of  couples of tuples that verifies the constraints expressed by all the gradual items of the pattern while Di-Jorio et al. \cite{Di-JorioLT09} define the support as the size of the longest sequence of tuples supporting a gradual pattern. More recently, Lonlac et al. \cite{Lonlac18} define the support of a gradual pattern respecting the temporal order as the proportion of couples of consecutive tuples supporting the gradual pattern.
In this paper, we follow this last definition of the support of a gradual pattern respecting the temporal order and we  propose a new one for discovering frequent seasonal gradual patterns.

To define the seasonal gradual itemsets mining problem, we introduce the following definitions:
\begin{definition}[Observations list respecting a gradual itemset]
\label{def:sequenceoftuples}
Let $g = (i_1^{*_1}, \ldots, i_k^{*_k})$ be a gradual k-itemset and
$L = (d_1, d_2, \ldots, d_n )$ be a list of consecutive observations.
We say that $L$ respects $g$ if for all $p \in [1, k]$, and for all $j \in [1, n-1]$, $d_j[i_p] *_p d_{j+1}[i_p]$ holds.
\end{definition}

Considering   database $\Delta_1$ from Table \ref{tab:exampleDataset2}, $L_1 = (d_1, d_2, d_3)$ is a list of consecutive observations with respect to $g_1 = \{age^{\uparrow},  salary^{\uparrow}\}$.
By definition, the size (cardinality) of a list of observations $|L|$ is the number of observations that it contains. Thus, the size of $L_1$ is equal to $3$.

\begin{definition}[Maximal observations list]
\label{def:maxmotif}
Let $g = (i_1^{*_1}, ... , i_k^{*_k})$ be a gradual itemset and let $L$ be a list of consecutive observations respecting $g$.
We say that $L$ is maximal if for any list of consecutive observations $L'$ respecting $g$, $L \not \subset L'$ and $|L| \geq |L'|$.
\end{definition}

For instance, in Table \ref{tab:exampleDataset2}, sequence $L_2 = (d_6, d_7, d_8, d_9)$ is not maximal with respect to $g_1$ because $L_3 = (d_5, d_6, d_7, d_8, d_9)$ is a list of consecutive observations with respect to $g_1$  and it contains $L_2$. $L_3$ is a maximal observations list with respect to $g_1$.

It is important to note that there may be several maximal lists of consecutive observations respecting $g$ for a single database. For instance $L_1$ and $L_3$ are two maximal lists of consecutive observations respecting $g_1$.

\begin{definition}[Cover]
\label{coverG}
Let $g$ be a gradual itemset extracted from a numerical database $\Delta$. We define $Cover(g, \Delta)$ as the set of maximal lists of tuples with respect to $g$ in $\Delta$.
\end{definition}

Considering   database $\Delta_1$ in Table \ref{tab:exampleDataset2} and the previous gradual itemset $g_1$,  $Cover(g_1, \Delta_1) =  \langle(d_1, d_2, d_3), (d_5, d_6, d_7, d_8, d_9) \rangle$.

Before defining the problem of mining seasonal gradual patterns, let us first describe the problem of finding periodic patterns in multiple sequences.

\subsection{Problem of finding periodic patterns in multiple sequences}
\label{PFPS}
Periodic patterns mining is one of the popular data mining tasks that has received considerable attention recently (e.g., \cite{Fournier-VigerL16,Fournier-VigerL19,AmphawanSL10}). As mentioned previously, periodic patterns can be used for marketing as they help to understand the purchase behavior of customers by discovering sets of items that are  bought periodically.
Many algorithms have been designed to find these patterns (e.g., \cite{Fournier-VigerL16,TanbeerAJL09,GalbrunCTTC18,Fournier-VigerL19}).
We describe in the following the problem of finding periodic pattern.

Let $\mathcal{I} = \{i_1, i_2,\dots, i_n\}$ be a set of items, where $n$ is the set cardinality. Let us assume that itemset $X$ is a subset of $\mathcal{I}$, that is $X \subseteq I$. A sequence $s$ is an ordered list of itemsets $s = \langle D_1, D_2,\dots, D_m\rangle$, where $D_j \subseteq \mathcal{I}$ ($1 \leq j\leq m$). 

$ s_1 = \langle (i_1,i_2,i_3), (i_2,i_4), (i_1,i_2,i_5), (i_3), (i_1,i_2,i_4,i_5), (i_1,i_3), (i_2,i_3), (i_2,i_5) \rangle$ is a sequence that contains eight itemsets. The first itemset contains three items ($i_1$, $i_2$ and $i_3$).

The concept of periods of an itemset was introduced in order to find periodic patterns in a single sequence \cite{inbook}. In the periodic patterns mining framework in multiple sequences, several other measures of periods were defined like the sequence periodic ratio \cite{Fournier-VigerL18,Fournier-VigerL19}.
In a retail store, analysing multiple sequences of customers' transactions is useful for example, to discover periodic patterns that are common to many customers.

\begin{definition}[Sequence]
\label{subsequence}
A sequence $s = \langle D_1, D_2,\dots, D_k\rangle$ is said to be a sub-sequence of a sequence $s' = \langle D'_1, D'_2,\dots, D'_l\rangle$, noted $s \subseteq s'$,  iff there exist integers $1 \leq a_1 < a_2 < \ldots < a_k \leq l$ such that $D_1 \subseteq D'_{a_1}$, $D_2 \subseteq D'_{a_2}$, $\ldots$, $D_k \subseteq D'_{a_k}$.
\end{definition}

For example $\langle (i_1,i_2), (i_1) \rangle$ is a sub-sequence of  sequence $s_1$.

\begin{definition}[List of consecutive transactions in a sequence]
\label{consecutiveTrans}
Let $s$ be a sequence and $X$ be an itemset. Let $TR(X,s) = \langle D_{a_1}, D_{a_2},\dots, D_{a_k} \rangle \subseteq s$ be an ordered set of transactions in which itemset $X$ occurs in sequence $s$.
Transactions $D_x$ and $D_y$ in $s$ are said to be consecutive with respect to $X$ if there does not exist a transaction $D_z \in s$ such that $x < z < y$ and $X \in D_z$.
\end{definition}

For example $(i_1, i_2, i_3)$ and $(i_1,  i_2, i_5)$ are two consecutive transactions with respect to the itemset $\{i_1, i_2\}$ in sequence $s_1$.

\begin{definition}[Period of an itemset in a sequence]
\label{period}
Let $s$ be a sequence and $X$ be an itemset. Let $TR(X,s) = \langle D_{a_1}, D_{a_2},\dots, D_{a_k}\rangle \subseteq s$ be an ordered set of transactions in which itemset $X$ occurs in sequence $s$. Let $D_x$ and $D_y$ be
two consecutive transactions in $s$ with respect to $X$. The period of $D_x$ and $D_y$ for $X$ is $per(D_x , D_y ) = y - x$.
The periods of $X$ in  sequence $s$ are $pr(X, s) = \{ per_1, per_2 , . . . , per_{k + 1} \}$ where $per_1 = a_1 - a_0$, $per_2 = a_2 - a_1$, $\ldots$, $per_{k + 1} = a_{k+1} - a_k$, and $a_0 = 0$ and  $a_{k+1} = n$, respectively, where $n$ is the number of transactions of  sequence $s$.
\end{definition}

The definition of the period of an itemset in a sequence satisfies the classical property of antimonotonicity allowing to reduce the search space during the mining process of periodic patterns.

For example, if we consider   sequence $s_1$,   itemset $\{i_1, i_2\}$ occurs in transactions $D_1, D_3, D_5$, thus  $TR(X,s_1) = \{D_1, D_3, D_5\}$ and the periods of itemset $\{i_1, i_2\}$ are $pr(\{i_1, i_2\} , s ) = \{1, 2, 2, 3\}$.

In a single sequence $s$, the periodicity of an itemset $X$ is usually assessed by using maximum periodicity measure \cite{Fournier-VigerL16} defined as $ maxPr(X, s) = argmax(pr(X, s))$.
The maximum periodicity of $\{i_1, i_2\}$ in $s_1$ is equal to $3$.

The \textbf{support of an itemset} $X$ in  sequence $s$ is the number of transactions containing $X$ in $s$, that is, $Support(X, s) = |TR(X, s)|$.
%
%
The Support of $\{i_1, i_2\}$ in $s_1$ is equal to $3$.

The problem of mining periodic patterns in a sequence $s$ is to find each itemset $X$ such that $Support(X, s)  \geq minSup$ and $maxPr(X, s) \leq maxPr$, where \texttt{minSup} is the minimum support threshold and \texttt{maxPr} is the maximum periodicity threshold.

To avoid finding periodic patterns with periods that vary too much in a sequence, a new periodicity measure based on the statistical measure standard deviation was considered \cite{Fournier-VigerL19}. Indeed, for an itemset $X$ in a sequence $s$, this measure is defined as the standard deviation of the periods of $X$ and is denoted $stanDev(X, s)$. Considering the previous sequence $s_1$, the periods of itemset $\{i_1, i_2\}$ in $s_1$ are $\{1, 2, 2, 3\}$, the average period is $(1+2+2+3)/4 = 2$ then $stanDev(\{i_1, i_2\}, s_1) = \sqrt{[(1-2)^2 + (2-2)^2 + (2-2)^2 + (3-2)^2]/4}$.


By considering a sequence database, the periodic pattern mining problem has been extended to discover periodic frequent patterns common to multiple sequences using new measures \cite{Fournier-VigerL19,Fournier-VigerL18}.
In the following, we give  a description of this problem.

\begin{definition}[Sequence database]
\label{SeqData}
A sequence database $\Delta$ is a set of $m$ sequences $\langle sid_i, s_i\rangle$ ($1\leq i\leq m$), denoted as $\Delta =\{(sid_1, s_1), \dots, (sid_m, s_m)\}$ ($1\leq i\leq m$), where $sid_i$ is a sequence identifier and $s_i$ a sequence.
\end{definition}


The cover of an itemset $X$ in a sequence database $\Delta$ is the set of sequence identifiers where the itemset is included i.e., $cover(X, \Delta) = \{sid_k ~|~ (sid_k, s_k) \in \Delta \wedge |TR(X, sid_k)| > 0 \}$.

Extracting periodic frequent patterns from a sequence database (multiple sequences) is performed using a measure called
the \textit{sequence periodic ratio} defined as follows:

\begin{definition}[Sequence periodic ratio]
\nllabel{def:stdP}
The sequence periodic ratio of an itemset $X$ in a sequence database $\Delta$ is defined as \\
$ra(X, \Delta) = \{s~|~ maxPer(X,s) \leq maxPr~\wedge~Support(X,s)~\geq~minSup \wedge \\ stanDev(X, s)~\leq~maxStd~\wedge s \in \Delta \}/|\Delta|$.
\end{definition}

For a given minimum sequence periodic ratio threshold $minRa$, an itemset $X$ is a periodic frequent pattern in a sequence database $\Delta$ if $ra(X, \Delta) \geq minRa$.

\begin{definition}[Periodic pattern mining problem in a sequence database]
\label{PeriodicPatternsMiningProblemInSequenceD}
Given a sequence database $\Delta$ and the following user-defined thresholds: the minimum support threshold \texttt{minSup}, the maximum periodicity threshold \texttt{maxPr}, the maximum standard deviation threshold \texttt{maxStd}, and the minimum sequence periodic ratio threshold \texttt{minRa}, the problem of mining periodic patterns in $\Delta$ is to find all periodic frequent patterns in $\Delta$ such that $Support(X, s) \geq minSup$ and $maxPr(X, s) \leq maxPr$ and $stanDev(X, s) \leq maxStd$ and $ra(X, \Delta) \geq minRa$.
\end{definition}

Recently, the algorithms named \texttt{MPFPS\_BFS} (Mining Periodic Frequent Pattern common to multiple Sequences using a breadth first search) and \texttt{MPFPS\_DFS} (using the depth first search)  were proposed in \cite{Fournier-VigerL19} to mine periodic frequent patterns in a sequence database.
These algorithms are based on two pruning properties and an effective data structure to reduce the search space.

\section{ Problem statement}
\label{problemStement}
In this section, we propose the problem of mining seasonal gradual patterns in a temporal data sequence and we study its properties. We describe the type of data used, the notion of seasonal gradual patterns, and its examples.

\subsection{Temporal data sequences}
Our approach finds its application on a numerical database $\Delta$ composed of temporal data sequences.
More precisely, the database $\Delta$ consists of sequences $\mathcal{S} = \langle s_1, s_2, \ldots, s_m \rangle$ described by the set of numerical attributes $\mathcal{I} = \{i_1, \ldots, i_n\}$,  with $s_j = (d_1, \ldots, d_l)$, a list of ordered periods (dates) considered.
Table \ref{tab:exampleseqDataset} is an example of temporal data sequences that gives information about customer purchases for a e-commerce website on three purchase cycles (sequences) $s_1, s_2, s_3$. Each sequence contains the data for eight periods ($d_1, \ldots, d_8$).
Without loss of generality, we assume that there are no  other purchase dates between two consecutive dates and that the purchases are made continuously between two consecutive cycles.

Contrary to most of  sequence patterns mining algorithms  proposed for mining patterns from a categorical sequence database, our proposed approach deals with a numerical sequence database.

\begin{table*}[htbp]
\centering
\begin{tiny}
\caption{Customer purchases database:  $\Delta_2$}
 \label{tab:exampleseqDataset}
\begin{tabular}{c|c|cccc}
 \hline
\textbf{Sid} & \textbf{purchase\_timestamp} & \textbf{age (a)} & \textbf{freight\_value (f)} & \textbf{payment\_installments (pi)} & \textbf{payment\_value (pv)} \\
\hline
 \multirow{8}{*}{$s_1$} & $d_1$ & \textbf{22} & 8.72 & \textbf{2} &  18.12 \\
& $d_2$ & \textbf{24} & 22.76 & \textbf{3}  & 141.46 \\
& $d_3$ & \textbf{28} & 19.22  & \textbf{4}  & 179.12 \\
& $d_4$ & 20 & 17.20 & 1  & 72.20 \\
& $d_5$ & 18 & 8.72 & 1  & 28.62\\
& $d_6$ & 35 & 27.36 & 3  & 175.26\\
& $d_7$ & 38 & 16.05 & 4  & 65.95\\
& $d_8$ & 44 & 15.17 & 4  & 75.16\\
\hdashline
 \multirow{8}{*}{$s_2$} & $d_1$ & \textbf{32} & 16.05 & \textbf{3} &  35.95 \\
& $d_2$ & \textbf{34} & 19.77 & \textbf{4}  & 161.42 \\
& $d_3$ & \textbf{36} & 30.53 & \textbf{5}  & 159.06 \\
& $d_4$ & \textbf{40} & 16.13 & \textbf{5}  & 114.13 \\
& $d_5$ & 25 & 14.23 & 2  & 50.13 \\
& $d_6$ & 23 & 12.805 & 2  & 32.70 \\
& $d_7$ & 20 & 13.11 & 1  & 54.36 \\
& $d_8$ & 41 & 14.05 & 4  & 46.45 \\
\hdashline
 \multirow{8}{*}{$s_3$} & $d_1$ & \textbf{28} & 77.45 & \textbf{3} &  1376.45 \\
& $d_2$ & \textbf{33} & 15.10 & \textbf{4}  & 43.09 \\
& $d_3$ & \textbf{38} & 11.85 & \textbf{6}  & 29.75 \\
& $d_4$ & 35 & 16.97 & 5  & 62.15 \\
& $d_5$ & 38 & 8.96 & 4  & 118.86 \\
& $d_6$ & 44 & 8.71 & 5  & 88.90 \\
& $d_7$ & 52 & 7.78 & 6  & 17.28 \\
& $d_8$ & 41 & 57.58 & 4  & 187.57 \\
\hline
\end{tabular}
\end{tiny}
\end{table*}

\subsection{Seasonal gradual patterns}

In the case of a single observation sequence, a gradual pattern corresponds to the one extracted by Lonlac et al.  \cite{Lonlac18}. However, in the seasonal gradual pattern context, we seek gradual patterns respected by the same list of observations.
To address this issue, we propose the definition of seasonal gradual patterns in which the notion of seasonality is introduced.
Let us consider the temporal data sequence in Table \ref{tab:exampleseqDataset} which one there are $3$ sequences. These data are extracted from a dataset regarding customer orders made at multiple marketplaces.

The goal is to extract  frequent co-variations between attribute values that occur frequently in identical periods, i.e., seasonal gradual patterns. We want to extract these patterns with the lists of periods associated that will represent the seasonality of each pattern.
To illustrate our approach, we start by introducing the following definitions.

\begin{definition}
\label{def:gradToSeq}
Let $\Delta$ be a temporal data sequence over a set of numerical attributes ${\mathcal I}= \{i_1, \ldots,  i_n\}$,  
and object (period) sequences $\mathcal{S} = \langle s_1, s_2, \ldots, s_m \rangle$.
Given gradual item $i^{*}$ with $i\in\mathcal{I}$, we define  $s^*_{i}$ as a sequence of ordered period lists  $\langle cover(i^*, s_1), cover(i^*, s_2), \ldots, cover(i^*, s_m)  \rangle$.
\end{definition}
Sequence of ordered period lists $s^*_{i}$ is composed of the maximal period lists that respect gradual item $i^*$ in each sequence.

Referring back to the example from Table \ref{tab:exampleseqDataset}, we have:
\begin{itemize}
    \item  $cover(age^{\uparrow}, s_1) = \langle (d_1, d_2, d_3), (d_5, d_6, d_7, d_8) \rangle$
     \item $cover(age^{\uparrow}, s_2) = \langle (d_1, d_2, d_3, d_4), (d_7, d_8) \rangle$
     \item $cover(age^{\uparrow}, s_3) = \langle (d_1, d_2, d_3), (d_4, d_5, d_6, d_7) \rangle$
\end{itemize}
Then $s_{age}^{\uparrow} = \langle (d_1, d_2, d_3), (d_5, d_6, d_7, d_8), (d_1, d_2, d_3, d_4), (d_7, d_8), (d_1, d_2, d_3), (d_4, d_5, d_6, d_7) \rangle$.

Note that a given gradual item $i^*$ corresponds to a unique maximal period lists sequence $s^*_{i}$.
%

\begin{definition}[Seasonal gradual item]
\label{seasonalgi}
Let $\Delta$ be a temporal data sequence over a set of numerical attributes ${\mathcal I}= \{i_1, \ldots,  i_n\}$ and $X$ an itemset consisting of the objects (periods) in $\Delta$. A seasonal gradual item is defined under the form of $i^{(*,X)}$, where $* \in \{\uparrow, \downarrow\}$.
\end{definition}

\begin{example}
\label{example2}
If we consider the temporal data sequence in Table \ref{tab:exampleseqDataset}, $age^{(\uparrow, \{d_1, d_2, d_3\})}$ is a seasonal gradual item expressing that  the values of  attribute $age$ are increasing  more frequently on the period $(d_1, d_2, d_3)$.
\end{example}

\begin{definition}[Seasonal gradual itemset (pattern)]
\label{seasonalgit}
A seasonal gradual itemset (SGP)  $g = \{i_{1}^{(*_{1},X)}, \ldots, i_{k}^{(*_{k}, X)}\}$ is a non-empty set of seasonal gradual items, with $X$ being an itemset consisting of the objects (periods) of $\Delta$.
\end{definition}

\begin{example}
\label{example3}
Consider the temporal data sequence in Table \ref{tab:exampleseqDataset}, \\
$g_1 = \{age^{(\uparrow, \{d_1, d_2, d_3\})}, \; payment\_installments^{(\uparrow, \{d_1, d_2, d_3\})}\}$ is a seasonal gradual pattern meaning that "an increase of $age$ comes along with an increase of  $payment\_installments$ more frequently in period $(d_1, d_2, d_3)$".
\end{example}

The support to measure  the graduality of seasonal gradual pattern is defined as follows:

\begin{definition}[Seasonal gradual itemset support computation]
Let $\Delta$ be a temporal data sequence and $g = \{i_{1}^{(*_{1},X)}, \ldots, i_{k}^{(*_{k}, X)}\}$ be a seasonal gradual itemset of $\Delta$. The support of $g$ in $\Delta$ can be defined as follows:
 \begin{equation}
     Support(g, \Delta) = \frac{\underset{1 \leq p \leq k}{min}~Support(X, s_{i_p}^{*_p}) }{|\Delta|}
 \end{equation}
 where $|\Delta|$ is the number of sequences in  $\Delta$, and $s_{i_p}^{*_p}$ the sequence of ordered period lists for gradual item $i_p^{*_p}$.
\end{definition}

\begin{example}
\label{example4}
Referring again to the temporal data sequence $\Delta_2$ in Table \ref{tab:exampleseqDataset} and the seasonal gradual itemset $g_1$, for $minSup = 2$, we have $Support(g_1, \Delta_2) = \frac{3}{3 }$ as $Support(\{d_1, d_2, d_3\}, s_{age}^{\uparrow}) = 3 $ and $Support(\{d_1, d_2, d_3\}, s_{pi}^{\uparrow}) = 3 $.
\end{example}

Given a predefined minimum support threshold $\theta$ and a seasonal gradual pattern $g$, we say that $g$ is frequent if its support is greater than or equal to $\theta$. The support definition of seasonal gradual pattern satisfies the classical anti-monotony property.

\begin{definition}[Seasonal gradual patterns mining problem]
\label{sgp}
Let $\Delta$ be a temporal data sequence and $\theta$ a minimal support threshold. The problem of mining seasonal gradual patterns is to find the set of all frequent seasonal gradual patterns of $\Delta$ with respect to $\theta$.
\end{definition}

Let us indicate that, in the classical periodic patterns mining framework, the problem of finding periodic patterns in a single sequence of items that has been extensively investigated in e.g., \cite{Fournier-VigerL16,GalbrunCTTC18} is related to the frequent seasonal gradual patterns mining problem.
In fact, for a gradual item $i^*$, mining all periodic frequent patterns of $S_i^{*}$ corresponds to the traditional problem of mining all maximal periodic patterns in a sequence of objects without maximum periodicity threshold constraint \textit{maxPr}.

\section{Extracting seasonal gradual patterns}
\label{SeasonalPatternsExtraction}

In this section, we describe how to extract seasonal gradual patterns from a numerical temporal data sequence. We first transform the frequent seasonal gradual patterns mining problem into the problem of finding all maximal periodic patterns common to multiple sequences.
This transformation is given by the following definition.

\begin{definition}
\label{trans_2}
Let $\Delta$ be a temporal data sequence over a set of numerical attributes $\mathcal{I} = \{i_1, \ldots, i_n\}$. We define the objects sequence database  $\Gamma(\Delta)$  associated to $\Delta$ as $\Gamma(\Delta) = \{(i_1^{\uparrow},S_{i_1}^{\uparrow}), (i_1^{\downarrow},S_{i_1}^{\downarrow}), \ldots, (i_n^{\uparrow},S_{i_n}^{\uparrow}), (i_n^{\downarrow},S_{i_n}^{\downarrow}) \}$.
\end{definition}

\begin{example}
\label{example5}
For instance, according to Definition \ref{trans_2}, the object sequences database associated to database $\Delta_2$ in Table \ref{tab:exampleseqDataset} is given in Table \ref{tab:transdatasetSeq}.
\end{example}


The notion of sequence periodic ratio ($ra$) of an itemset was defined in \cite{Fournier-VigerL19} for mining periodic patterns common to multiple sequences (finding periodic patterns in a sequence database).
In order to exploit frequent periodic patterns mining algorithms for extracting seasonal gradual patterns, we modify the sequence periodic ratio definition as follows:


\begin{definition}
\label{ratioPeSeq}
Let $\Delta$ be a temporal data sequence and $g = \{i_{1}^{(*_{1},X)}, \ldots, i_{k}^{(*_{k},X)}\}$ a seasonal gradual pattern of $\Delta$. We define the sequence periodic ratio of $X$ in $\Gamma(\Delta)$  as  $ra(X, \Gamma(\Delta)) = \frac{|\{s ~|~ Support(X, s) ~\geq ~minSupp ~\wedge~  s~ \in ~\Gamma(\Delta)\}|}{|\Gamma(\Delta)|}$.
\end{definition}

The sequence periodic ratio definition of a pattern given in this paper is different of that given by \cite{Fournier-VigerL19}. In fact, the proposed  sequence periodic ratio definition (see definition \ref{ratioPeSeq}) for extracting seasonality satisfies the classic anti-monotony property, thus allowing to efficiently extract frequent seasonal gradual patterns.

\begin{propriete}
\label{Property1}
Let $\Delta$ be a temporal data sequence and $g = \{i_{1}^{(*_{1},X)}, \ldots, i_{k}^{(*_{k},X)}\}$ a frequent seasonal gradual pattern extracted from $\Delta$, then $ra(X, \Gamma(\Delta)) \geq k \times |\Gamma(\Delta)|$.
\end{propriete}

As we find the seasonal gradual patterns $g = \{i_{1}^{(*_{1},X)}, \ldots, i_{k}^{(*_{k},X)}\}$ from $\Delta$ of size $k \geq 1$, we calculate the minimum sequence periodic ratio threshold \texttt{minRa} equal to $1/|\Gamma(\Delta)|$ to extract frequent periodic patterns from $\Gamma(\Delta)$ corresponding to the seasonality of our seasonal gradual patterns.



\begin{table}[htbp]
\centering
\begin{scriptsize}
\renewcommand{\arraystretch}{1.5}
\caption{Tuple sequences database $\Gamma(\Delta_2)$ obtained from database $\Delta_2$ }
 \label{tab:transdatasetSeq}
\begin{tabular}{cl}
 \hline
\textbf{Gradual items} & \textbf{Tuple sequences}  \\
\hline
$a^\uparrow$ & $\langle (d_1, d_2, d_3), (d_5, d_6, d_7, d_8), (d_1, d_2, d_3, d_4), (d_7, d_8), (d_1, d_2, d_3), (d_4, d_5, d_6, d_7) \rangle$ \\
$a^\downarrow$ &  $\langle (d_3, d_4, d_5), (d_8, d_1), (d_4, d_5, d_6, d_7), (d_8, d_1),  (d_3, d_4), (d_7, d_8) \rangle$ \\
$f^\uparrow$ &  $ \langle( d_1, d_2) (d_5, d_6), (d_8, d_1, d_2, d_3), (d_6, d_7, d_8, d_1), (d_3, d_4), (d_7, d_8) \rangle$ \\
$f^\downarrow$ & $\langle (d_2, d_3, d_4, d_5), (d_6, d_7, d_8), (d_3, d_4), (d_5, d_6), (d_1, d_2, d_3), (d_4, d_5, d_6, d_7) \rangle$ \\
$pi^\uparrow$ &  $\langle (d_1, d_2, d_3), (d_4, d_5, d_6, d_7, d_8), (d_1, d_2, d_3, d_4), (d_5, d_6), (d_7, d_8), (d_1, d_2, d_3), (d_5, d_6, d_7) \rangle$ \\
$pi^\downarrow$ & $\langle (d_3, d_4, d_5), (d_7, d_8, d_1), (d_3, d_4, d_5, d_6, d_7), (d_8, d_1),  (d_3, d_4, d_5), (d_7, d_8) \rangle$ \\
$pv^\uparrow$ & $\langle (d_1, d_2, d_3), (d_5, d_6), (d_7, d_8), (d_1, d_2), (d_6, d_7), (d_8, d_1), (d_3, d_4, d_5), (d_7, d_8) \rangle$  \\
$pv^\downarrow$ & $\langle (d_3, d_4, d_5), (d_6, d_7), (d_8, d_1), (d_2, d_3, d_4, d_5, d_6),  (d_7, d_8), (d_1, d_2, d_3), (d_5, d_6, d_7) \rangle$ \\
\hline
\end{tabular}
\end{scriptsize}
\end{table}

Proposition \ref{prop1_1} illustrates the mapping between the set of seasonal gradual itemset  of $\Delta_2$ and the periodic patterns of $\Gamma(\Delta_2)$.

\begin{proposition}
\label{prop1_1}
Let $\Delta$ be a temporal data sequence and $\theta$ a minimal support threshold. $g = \{i_{1}^{(*_{1},X)}, \ldots, i_{k}^{(*_{k},X)}\}$ is a frequent seasonal gradual pattern of $\Delta$ if
$\forall 1 \leq p \leq k$,  $Support(X, S_{i_p}^{*_p}) > \theta \times |\Delta|$, with $Cover(X, \Gamma(\Delta)) = g$. Moreover $Cover(g, \Delta)$ is a subset of periodic frequent patterns $X$  of  $\Gamma(\Delta)$  with respect to a minimum sequence periodic ratio threshold \texttt{minRa} and with $Cover(X, \Gamma(\Delta)) = g$.
\end{proposition}


\begin{preuve}
\label{Proof-1}
According to our transformation, it is clear that $g = \{i_{1}^{(*_{1},X)}, \ldots, i_{k}^{(*_{k},X)}\}$ is frequent  if the cover of $X$ over sequences of $\Gamma(\Delta)$ consists of identifiers $i_1^{*_1} \ldots i_k^{*_k}$.
As the cover of $g$ contains the set of maximal object lists of $\Delta$ satisfying $g$, it is easy to see that the sequence periodic ratio definition of $X$ in $\Gamma(\Delta)$ allows to guarantee  the frequency of $g$ in $\Delta$.
\end{preuve}


\begin{propriete}
\label{Property2}
If $X$ is a periodic frequent pattern of $\Gamma(\Delta)$ then $g = cover(X, \Gamma(\Delta))$ is a  seasonal gradual pattern of $\Delta$.
\end{propriete}


%

After reducing the frequent seasonal gradual patterns mining problem from a temporal numerical data sequence to a problem of mining frequent periodic patterns in a sequence database as illustrated by Table \ref{tab:transdatasetSeq},  we use a modified version of \textit{MPFPS\_BFS} (Mining Periodic Frequent Pattern common to multiple Sequences)  algorithm \cite{Fournier-VigerL19} named here \textit{MPFPS\_BFS\_Modified}  on the obtained object lists sequence database to extract frequent periodic patterns which correspond to   seasonality. Then, for each extracted periodic pattern, the set of sequence identifiers containing the pattern forms a seasonal gradual pattern.
The modified version of \textit{MPFPS\_BFS} is given by Algorithm \ref{alg:MPFPS_Modif}.
\textit{MPFPS\_BFS\_modified}  algorithm performs the breadth-first search to discover all periodic frequent patterns common to multiple sequences for a given sequence database.
It takes as input a database with multiple sequences and a minimum support threshold, and  outputs all periodic frequent patterns common to multiple sequences.

It is worth mentioning that state-of-the art gradual patterns extraction algorithms would fail to find seasonal gradual patterns as the object sequences respecting two gradual items of a gradual pattern can belong to two different sequences. For instance, seasonal gradual pattern $\{i_{1}^{(*_{1},X)}, i_{2}^{(*_{2},X)}\}$, where itemset $X$ respects $i_1^{*_1}$ in sequence $s_1$ and does not respect $i_2^{*_2}$ in sequence $s_1$ but rather in sequence $s_2$ cannot be found using state-of-the art gradual patterns extraction algorithms.
These traditional algorithms for gradual pattern mining cannot be directly applied to the proposed problem because they were designed to handle a single sequence.


\begin{lemme}
\label{lem1}
Let $\Delta$ be a temporal data sequence over an attribute set $\mathcal{I}$, and let $s_i^*$ ($i \in \mathcal{I}$) be a sequence of ordered object lists extracted from  $\Delta$. Then $|s_i^*| \leq |\Delta|$.
\end{lemme}




In Algorithm \ref{alg:MPFPS_Modif}, the \textit{boundRa} measure is identical to \textit{ra} measure. Moreover, \textit{ra} measure   used in Algorithm \ref{alg:MPFPS_Modif} is anti-monotone and thus can be directly used to reduce the search space. This is not the case for the \textit{ra} measure used  in Algorithm \ref{alg:MPFPS}. Let us recall that the \textit{boundRa} measure is a an upper-bound on the \textit{ra} measure.
Algorithm \ref{alg:MPFPS_Modif} does not use \textit{maxStd, maxPr} measure as   Algorithm \ref{alg:MPFPS} since these measures are not necessary to compute our seasonal gradual itemsets.

\begin{algorithm}[!h]
  \caption{\textit{The  MPFPS\_BFS algorithm \cite{Fournier-VigerL19}}}
  \label{alg:MPFPS}
  \SetInd{0.3em}{0.7em}
  \KwIn{ $\Delta$ : a sequence database, a database with multiple sequences, maxStd, minRa, maxPr, minSup: the thresholds.}
  \KwOut{ : the set of periodic frequent patterns (PFPS).}
  \SetKwBlock{Deb}{Begin}{End}
   {
   Scan each sequence $s \in \Delta$ to calculate Support(\{i\}, s), pr(i, s), maxPr(i, s) and stanDev(\{i\}, s) for each item  $i \in \mathcal{I}$\;
   \For {\textbf{all}  item $i \in \mathcal{I}$}
     {
     $numSeq(\{i\}) \longleftarrow |\{ s|~maxpr(\{i\}, s\}) \leq maxPr \wedge stanDev(\{i\}, s) \leq maxStd \wedge Support(\{i\}, s) \geq minSup \wedge s \in \Delta|$ \;
     $ra(\{i\}) \longleftarrow \frac{numSeq(\{i\})}{|\Delta|}$ \;
    \If{$ra(\{i\}) \geq minRa$} {output $\{i\}$ \;}
    $numcand(\{i\}) \longleftarrow |\{ s|~maxpr(\{i\}, s\}) \leq maxPr \wedge Support(\{i\}, s) \geq minSup \wedge s \in \Delta|$ \;
    $boundRa(\{i\}) \longleftarrow \frac{numCand(\{i\})}{|\Delta|}$ \;
    }
    $boundPFPS \longleftarrow \{PFPS-list ~ of ~ item ~ i | i \in \mathcal{I} \wedge boundRa(\{i\}) \geq minRa \}$
   \While{$|boundPFPS| \geq 2$}
    {
     $GenerateItemsets(boundPFPS, minSup, maxPr, maxStd, minRa, \Delta)$
    }
  }
\end{algorithm}

\begin{algorithm}[!h]
  \caption{\textit{The  MPFPS\_BFS\_modified algorithm }}
  \label{alg:MPFPS_Modif}
  \SetInd{0.3em}{0.7em}
  \KwIn{ $\Delta$ : a sequence database, minRa, minSup: the thresholds.}
  \KwOut{ : the set of periodic frequent patterns (PFPS) with sequence identifiers containing them.}
  \SetKwBlock{Deb}{Begin}{End}
   {
   Scan each sequence $s \in \Delta$ to calculate Support(\{i\}, s) for each item  $i \in \mathcal{I}$\;
   \For {\textbf{all}  item $i \in \mathcal{I}$}
     {
     $numSeq(\{i\}) \longleftarrow |\{ s|~ Support(\{i\}, s) \geq minSup \wedge s \in \Delta|$ \;
    $ra(\{i\}) \longleftarrow \frac{numSeq(\{i\})}{|\Delta|}$ \;
    \If{$ra(\{i\}) \geq minRa$}
     {
        output $\{i\}$ \;
        output $numSeq(\{i\})$ \;
         $boundPFPS \longleftarrow~ boundPFPS ~\cup~ ~\{PFPS-list ~ of ~ item ~ i \}$ \;
     }
    }
   \While{$|boundPFPS| \geq 2$}
    {
     $GenerateItemsets(boundPFPS, minSup, minRa, \Delta)$
    }
  }
\end{algorithm}

\begin{algorithm}[!h]
  \caption{\textit{The  \textit{MSGP} algorithm }}
  \label{alg:dp}
  \SetInd{0.3em}{0.7em}
  \KwIn{ $\Delta$ : a temporal data sequence, $\theta$ : a minimum support threshold.}
  \KwOut{$\delta$ : the set of frequent seasonal gradual patterns (FSGP).}
  \SetKwBlock{Deb}{Begin}{End}
   {
   $\delta \longleftarrow \emptyset$  \;
   $\Gamma(\Delta) \longleftarrow \emptyset$  \tcc*{the tuples sequence database}
    \For {\textbf{all} gradual item $i^* (i \in \mathcal{I} \wedge * \in \{\uparrow, \downarrow\})$}
    {   $S_{i}^* \longleftarrow \emptyset$ \;
        \For {\textbf{all} sequence $s \in \Delta$}
        {
            $S_{i}^* \longleftarrow S_{i}^* \cup cover(i^*, s)$
        }
        $\Gamma(\Delta) \longleftarrow \Gamma \cup (i^*, S_{i}^*)$ \;
    }
    $SI \longleftarrow MPFPS\_BFS\_Modified(\Gamma(\Delta),1/|\Gamma(\Delta)|, \theta)$ \tcc*{the set of seasonality with sequence identifiers containing them}

    \For {\textbf{all} $si \in SI$}
    {
       $g \longleftarrow \emptyset$ \;
        \For {\textbf{all} $(sea, i^*) \in si$}
        {
            $g \longleftarrow g \cup i^{(*,sea)}$
        }
        \If{$g \not \subset \delta$}
        {
            $\delta \longleftarrow \delta \cup g$ \;
        }
    }
    \Return $\delta$ \;

  }
\end{algorithm}


\subsection*{Complexity, Correctness and completeness of the MSGP algorithm}

The time complexity of the  \textit{MSGP} algorithm depends on the time complexity of the \textit{MPFPS\_BFS\_modified} algorithm and the polynomial-time complexity to reduce seasonal gradual patterns mining problem to the problem of mining frequent periodic patterns in a sequence database.
The overall computational time cost of the \textit{MPFPS\_BFS\_modified} algorithm is $O(n \times w + |I| )$ to evaluate the 1-itemsets, and then $O(n \times w )$ for each itemset considered in the search space, where $w$ is the average number of transactions per sequence and $n$ the number of sequences \cite{Fournier-VigerL19}.

Algorithm \textit{MSGP} is correct and complete for frequent seasonal gradual itemset mining since it depends on \textit{MPFPS\_BFS} algorithm that has been shown correct and complete to compute all frequent periodic patterns in a sequence database \cite{Fournier-VigerL19}.

\section{Experimental evaluation}
\label{Experiments}

The major interest of seasonal gradual patterns is that they are well adapted to capture some common co-variations repeated with identical periods on attributes in the ordered data set. One such kind of data that receives a lot of attention nowadays is temporal data, i.e. data produced with a temporal order on the objects, often in e-commerce domain.
In order to illustrate the proposed method and show its performance, an experiment study has been conducted on real-world data sets.
We used a real-world data set of Federal Reserve's time series of foreign exchange rates per dollar from 2000 to 2019 taken from  \textit{Kaggle\footnote{https://www.kaggle.com/brunotly/foreign-exchange-rates-per-dollar-20002019}} and  a real-world data set taken from the \texttt{UCI Machine Learning Repository\footnote{https://archive.ics.uci.edu/ml/datasets.php}},  which is the Stock Exchange data, in order to find out relevant knowledge (e.g., seasons to invest) and give potentially useful suggestions to people who intend to invest.
We also carried out an experiment on a synthetic data set generated using IBM Synthetic Data Generation Code for Associations and Sequential Patterns.

The Foreign Exchange Rates data set contains $5217$ lines on $22$ attributes.
The lines of this data set represent the exchange rates of different currencies with the US dollar and the attributes represent the different currencies (items) with an attribute date indicating the dates of correspondence of each
rate with the US dollar.
The values of attribute date given over the format ``aaaa-mm-dd'' are ordered  and represent items of the sequences database.
For our experiments, we retrieve the order days from the order date attribute and consider them as temporal variables ($d_1, \ldots, d_m$).
The foreign exchange rates data set contains $3.7\%$ of missing data; we removed all the lines with missing data and the result data set contains $5019$ lines on $22$ attributes.

The stock exchange data set is collected from \texttt{imkb.gov.tr} and \texttt{finance.yahoo.com.} and  is organized with regards to working days in Istanbul Stock Exchange.
Its lines represent returns of Istanbul Stock Exchange (\textit{ISE}), with seven other international indices; Standard \& poor$\hat{a}$\euro{}$^{TM}$S 500 return index (SP), Stock market return index of Germany (\textit{DAX}), Stock market return index of the UK (\textit{FTSE}), Stock market return index of Japan (\textit{NIKKEI}), Stock market return index of Brazil (\textit{BOVESPA}), MSCI European index (\textit{MSCE\_EU}), MSCI emerging markets index (\textit{MSCI\_EM}) with regards to working days from June $5$, $2009$ to February $22$, $2011$.
The first column stands for the date  from June $5$, $2009$ to February $22$, $2011$ and all next columns of this data set stand for Istanbul Stock Exchange with other international index.
This data set is therefore suitable to our study framework, there is a temporal constraint on the data as the data lines are ordered by ascending year. Each year is ordered by ascending months and each month by ascending days. It contains $536$ lines and $9$ attributes which are stock market return index. The different days constitute the temporal variables $(d_1, \ldots, d_m)$.

The proposed algorithm \textit{MSGP} takes two parameters as input: \textit{minSup}, the minimum support threshold used to find frequent periodic patterns in each sequence   and \textit{minRa}, the minimum sequence periodic ratio threshold used to find frequent periodic patterns common to many sequences (the seasonality of the seasonal gradual patterns).

The \textit{MSGP} algorithm is implemented in Java, and it was run on a 2.8GHz Intel Core i7 CPU, 32GB memory with 8 cores.
The \textit{MPFPS\_BFS} algorithm was programmed in Java, we extend it to obtain our \textit{MSGP} algorithm.
We assess the performance of our algorithm in terms of runtime, number of patterns found and scalability, when varying the support. We compare the performance of the \textit{MSGP} algorithm with the current state of the art gradual patterns mining algorithms \cite{Lonlac18} called here ``TemporalGradual'' and \textit{Paraminer} \cite{NegrevergneTRM14} as the latter compute all classical gradual patterns and do not eliminate nonseasonal patterns. We show that  the \textit{MSGP} algorithm is able to filter many nonseasonal gradual patterns and extract a small number of patterns that are easy for a user to interpret.
To the best of our knowledge, there exists no algorithm (before the \textit{MSGP}) for mining seasonal frequent gradual patterns.
The proposed algorithms take  as input two parameters:  \textit{minSup}, the minimum support threshold used to find frequent periodic patterns in each sequence and \textit{minRa}.

\begin{figure}[!h]
\begin{center}
\includegraphics[width=2.3in,angle=270]{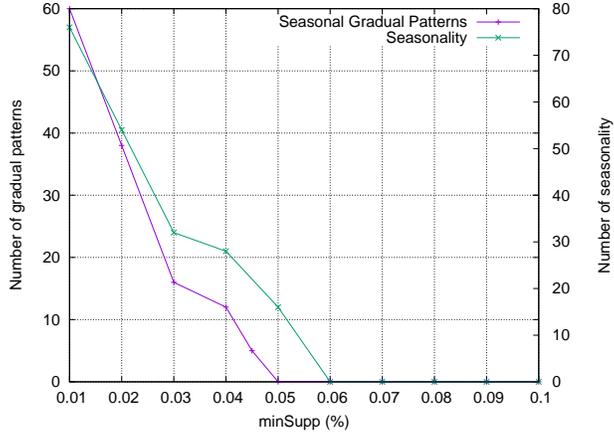}
  \caption{\bf Comparative evaluation in number: seasonal gradual patterns vs   seasonality (on a real-world data set of returns of Stock Exchange).}
  \label{fig:seasonalGradualPatterns}
  \end{center}
\end{figure}

Figure \ref{fig:seasonalGradualPatterns} shows the number of seasonal gradual patterns found by the \textit{MSGP} algorithm for various values of minimum support threshold  \textit{minSup} from The Stock Exchange data set.
One can remark from Figure \ref{fig:seasonalGradualPatterns} that the number of seasonal gradual patterns and the number of seasonality decrease  when the minimum support threshold increases.
The number of seasonality is always greater than the number of seasonal gradual patterns when the minimum support threshold increases as a seasonal pattern can be associated with more than one seasonality.

\begin{figure}[!h]
\begin{center}
\includegraphics[width=2.3in,angle=270]{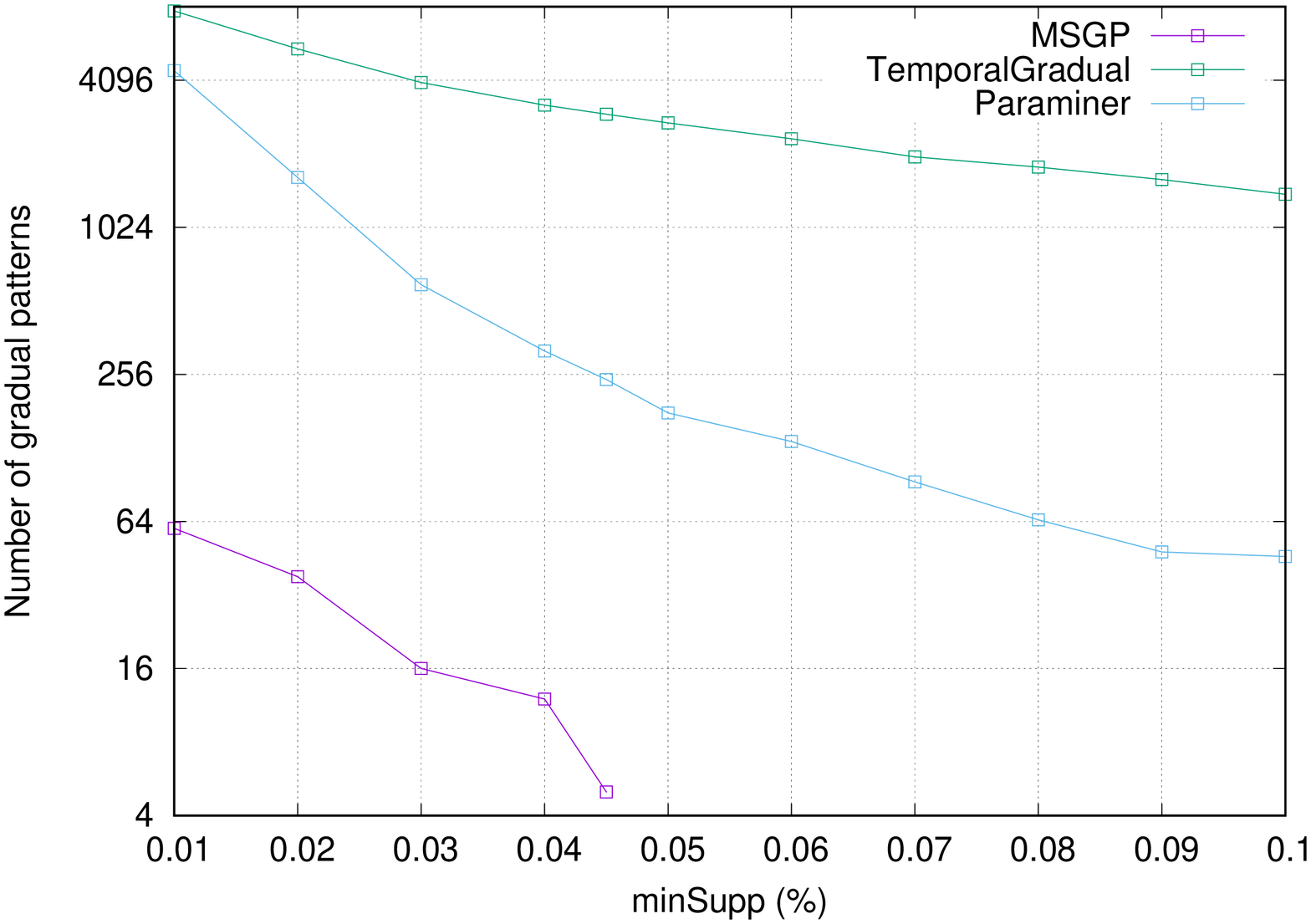}
  \caption{\bf Comparative evaluation in number: seasonal gradual patterns vs   classical gradual patterns (on a real-world data set of returns of Stock Exchange).}
  \label{fig:seasonalGradualPatterns-Compare}
  \end{center}
\end{figure}

Figure \ref{fig:seasonalGradualPatterns-Compare} compares the number of extracted patterns by the \textit{MSGP} algorithm with the number of temporal gradual patterns (TemporalGradual) and the number of patterns extracted with \textit{Paraminer} algorithm for various values of the minimum support threshold,  \textit{minSup}.
One can observe that the number of seasonal patterns is always small compared to other algorithms. This shows that the \textit{MSGP} algorithm can filter many nonseasonal gradual patterns to extract a reasonable quantity of seasonal gradual patterns with their associated seasonality what is easy to manage for a domain expert.

\begin{figure}[!h]
\begin{center}
\includegraphics[width=2.3in,angle=270]{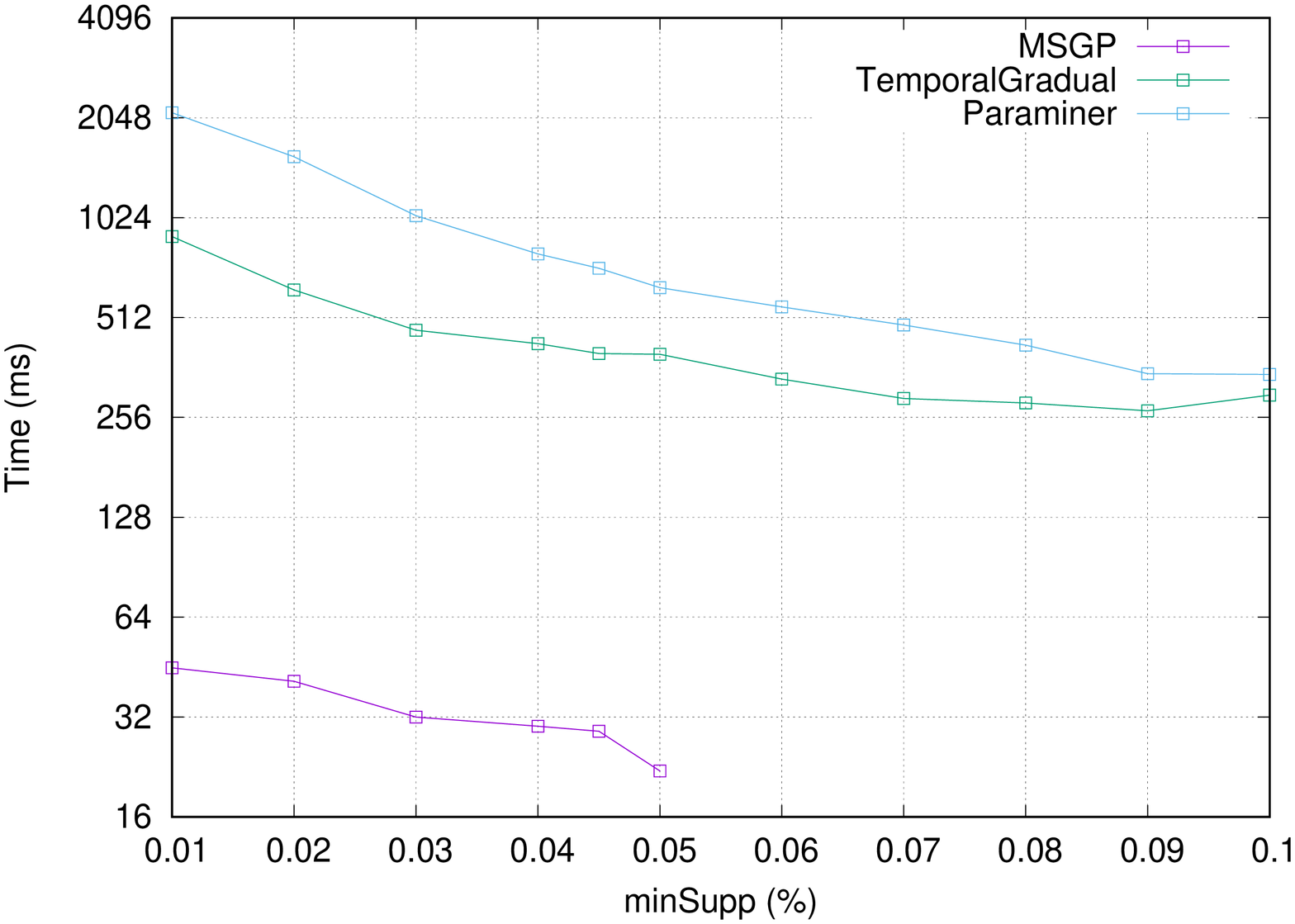}
  \caption{\bf Comparative evaluation in run-time: seasonal gradual patterns vs   classical gradual patterns (on a real-world data set of returns of Stock Exchange).}
  \label{fig:seasonalGradualPatterns-compraeTime}
  \end{center}
\end{figure}

Figure \ref{fig:seasonalGradualPatterns-compraeTime} shows the computational time taken by the different algorithms when varying the minimum support threshold  \textit{minSup}, with a logarithmic scale for time.
The \textit{MSGP} algorithm exhibits a better speedup comparatively to \textit{Paraminer} and \textit{TemporalGradual} algorithms.
The runtimes of \textit{MSGP} algorithm are very small as it extracts fewer patterns.

\section{Conclusion and perspectives}
\label{conclusion}
In this paper, we proposed an approach  to extract seasonal correlations between attributes from a graduality point of view in a temporal data sequence.
Seasonal gradual patterns mining problem is formulated as a problem of mining periodic frequent patterns common to multiple sequences.
The \textit{MSGP} (Mining Seasonal Gradual Patterns) algorithm was proposed to efficiently find all seasonal gradual patterns.   
We also proposed a definition of the associated support measure at a seasonal gradual pattern to efficiently mine frequent patterns in temporal data sequence  context.
The experimental evaluation on the e-commerce real world data shows that our approach is efficient and is of practical importance in the real world context.
In the future work, we will enrich the experimental study and  check the applicability of the approach to other temporal data, e.g., the data regarding the flow of product stocks. Seasonal gradual patterns extracted from such data will allow data experts to detect seasonal co-variations between quantity of products for better management of the supply chain.
Furthermore, we plan to extend the model to discover seasonal gradual rules for their application in recommendation systems.



\bibliography{biblio}

\begin{thebibliography}{10}
\expandafter\ifx\csname url\endcsname\relax
  \def\url#1{\texttt{#1}}\fi
\expandafter\ifx\csname urlprefix\endcsname\relax\def\urlprefix{URL }\fi
\expandafter\ifx\csname href\endcsname\relax
  \def\href#1#2{#2} \def\path#1{#1}\fi

\bibitem{Srikant1996J}
S.~Ramakrishnan, A.~Rakesh, Mining quantitative association rules in large
  relational tables, SIGMOD Rec. 25~(2) (1996) 1--12.

\bibitem{AumannL99}
Y.~Aumann, Y.~Lindell, A statistical theory for quantitative association rules,
  in: {SIGKDD}, 1999, pp. 261--270.

\bibitem{Salleb-AouissiVN07}
A.~Salleb{-}Aouissi, C.~Vrain, C.~Nortet, Quantminer: {A} genetic algorithm for
  mining quantitative association rules, in: {IJCAI}, 2007, pp. 1035--1040.

\bibitem{KaytoueKN11}
M.~Kaytoue, S.~O. Kuznetsov, A.~Napoli, Revisiting numerical pattern mining
  with formal concept analysis, in: {IJCAI}, 2011, pp. 1342--1347.

\bibitem{american2013diagnostic}
A.~P. Association, et~al., Diagnostic and statistical manual of mental
  disorders, BMC Med 17 (2013) 133--137.

\bibitem{AgierPS07}
M.~Agier, J.~Petit, E.~Suzuki, Unifying framework for rule semantics:
  Application to gene expression data, Fundam. Inform. 78~(4) (2007) 543--559.

\bibitem{BringayLOST09}
S.~Bringay, A.~Laurent, B.~Orsetti, P.~Salle, M.~Teisseire, Handling fuzzy gaps
  in sequential patterns: Application to health, in: {FUZZ-IEEE}, International
  Conference on Fuzzy Systems, {IEEE}, 2009, pp. 1338--1345.

\bibitem{DoTLNTA15}
T.~D.~T. Do, A.~Termier, A.~Laurent, B.~N{\'{e}}grevergne, B.~O. Tehrani,
  S.~Amer{-}Yahia, {PGLCM:} efficient parallel mining of closed frequent
  gradual itemsets, KAIS 43~(3) (2015) 497--527.

\bibitem{Hullermeier02}
E.~H{\"{u}}llermeier, Association rules for expressing gradual dependencies,
  in: {PKDD}, 2002, pp. 200--211.

\bibitem{BerzalCSVS07}
F.~Berzal, J.~C. Cubero, D.~S{\'{a}}nchez, M.~A.~V. Miranda, J.~Serrano, An
  alternative approach to discover gradual dependencies, IJUFKS 15~(5) (2007)
  559--570.

\bibitem{OudniLR13}
A.~Oudni, M.~Lesot, M.~Rifqi, Processing contradiction in gradual itemset
  extraction, in: {FUZZ-IEEE}, 2013, pp. 1--8.

\bibitem{Di-JorioLT09}
L.~Di-Jorio, A.~Laurent, M.~Teisseire, Mining frequent gradual itemsets from
  large databases, in: {IDA}, 2009, pp. 297--308.

\bibitem{NinLP10}
J.~Nin, A.~Laurent, P.~Poncelet, Speed up gradual rule mining from stream data!
  {A} b-tree and owa-based approach, J. Intell. Inf. Syst. 35~(3) (2010)
  447--463.

\bibitem{NegrevergneTRM14}
B.~N{\'{e}}grevergne, A.~Termier, M.~Rousset, J.~M{\'{e}}haut, Para miner: a
  generic pattern mining algorithm for multi-core architectures, DMKD 28~(3)
  (2014) 593--633.

\bibitem{PhanIMPT15}
N.~Phan, D.~Ienco, D.~Malerba, P.~Poncelet, M.~Teisseire, Mining
  multi-relational gradual patterns, in: SDM, 2015, pp. 846--854.

\bibitem{Masseglia08gradualtrends}
F.~Masseglia, A.~Laurent, M.~Teisseire, Gradual trends in fuzzy sequential
  patterns, in: IPMU, 2008, pp. 456--463.

\bibitem{Lonlac18}
J.~Lonlac, Y.~Miras, A.~Beauger, V.~Mazenod, J.-L. Peiry, E.~Mephu, An approach
  for extracting frequent (closed) gradual patterns under temporal constraint,
  in: {FUZZ-IEEE}, 2018, pp. 878--885.

\bibitem{OwuorLO19}
D.~Owuor, A.~Laurent, J.~Orero, Mining fuzzy-temporal gradual patterns, in:
  FUZZ-IEEE, 2019, pp. 1--6.

\bibitem{JabbourLS19}
S.~Jabbour, J.~Lonlac, L.~Sa{\"{\i}}s, Mining gradual itemsets using sequential
  pattern mining, in: {FUZZ-IEEE}, 2019, pp. 138--143.

\bibitem{Fournier-VigerL16}
P.~Fournier{-}Viger, J.~C. Lin, Q.~Duong, T.~Dam, {PHM:} mining periodic
  high-utility itemsets, in: {ICDM}, 2016, pp. 64--79.

\bibitem{TanbeerAJL09}
S.~K. Tanbeer, C.~F. Ahmed, B.~Jeong, Y.~Lee, Discovering periodic-frequent
  patterns in transactional databases, in: {PAKDD}, 2009, pp. 242--253.

\bibitem{GalbrunCTTC18}
E.~Galbrun, P.~Cellier, N.~Tatti, A.~Termier, B.~Cr{\'{e}}milleux, Mining
  periodic patterns with a {MDL} criterion, in: {ECML} {PKDD}, 2018, pp.
  535--551.

\bibitem{Fournier-VigerL19}
P.~Fournier{-}Viger, Z.~Li, J.~C. Lin, R.~U. Kiran, H.~Fujita, Efficient
  algorithms to identify periodic patterns in multiple sequences, Inf. Sci. 489
  (2019) 205--226.

\bibitem{Fong11}
A.~Fong, B.~Zhou, S.~Hui, G.~Hong, Web content recommender system based on
  consumer behavior modeling, IEEE Transactions on Consumer Electronics - IEEE
  TRANS CONSUM ELECTRON 57 (2011) 962--969.

\bibitem{lonlacIDA2020}
J.~Lonlac, E.~Mephu~Nguifo, A novel algorithm for searching frequent gradual
  patterns from an ordered data set, Vol.~24, 2020. to appear.

\bibitem{CaldersGJ06}
T.~Calders, B.~Goethals, S.~Jaroszewicz, Mining rank-correlated sets of
  numerical attributes, in: KDD, 2006, pp. 96--105.

\bibitem{AmphawanSL10}
K.~Amphawan, A.~Surarerks, P.~Lenca, Mining periodic-frequent itemsets with
  approximate periodicity using interval transaction-ids list tree, in: {WKDD},
  2010, pp. 245--248.

\bibitem{inbook}
P.~Fournier~Viger, C.-W. Lin, Q.-H. Duong, T.-L. Dam, L.~Sevcik, D.~Uhrin,
  M.~Vozňák, PFPM: Discovering Periodic Frequent Patterns with Novel
  Periodicity Measures, 2017.

\bibitem{Fournier-VigerL18}
P.~Fournier{-}Viger, Z.~Li, J.~C. Lin, R.~U. Kiran, H.~Fujita, Discovering
  periodic patterns common to multiple sequences, in: DaWaK, 2018, pp.
  231--246.

\end{thebibliography}

\end{document}